\documentclass{article} 
\usepackage{iclr2026_conference,times}

\usepackage{amsmath,amsfonts,bm}

\def\eqref#1{equation~\ref{#1}}

\def\1{\bm{1}}

\DeclareMathAlphabet{\mathsfit}{\encodingdefault}{\sfdefault}{m}{sl}
\SetMathAlphabet{\mathsfit}{bold}{\encodingdefault}{\sfdefault}{bx}{n}

\usepackage{hyperref}
\usepackage{url}
\usepackage{booktabs}
\usepackage{booktabs}
\usepackage{tabularx}
\usepackage{fancyvrb}
\usepackage{array}
\usepackage{longtable}
\usepackage{dirtree}
\usepackage[table]{xcolor}
\usepackage{array}
\usepackage{xcolor}
\usepackage{xspace}
\usepackage{multirow}
\usepackage{graphicx}

\hypersetup{colorlinks=true, citecolor=blue, linkcolor=blue}

\definecolor{nblue}{cmyk}{0.95,0.0,0.2,0.2}

\newcommand{\method}{\texttt{MLE-Smith}\xspace}
\RecustomVerbatimCommand{\VerbatimInput}{VerbatimInput}{fontsize=\footnotesize,
 frame=single,  
 framesep=0.5em, 
 labelposition=topline,
}

\title{MLE-Smith: Scaling MLE Tasks with Automated Multi-Agent Pipeline}

\iclrfinalcopy

\author{\textbf{Rushi Qiang}$^{1}$, \textbf{Yuchen Zhuang}$^{1}$, \textbf{Anikait Singh}$^{2}$ \\
\textbf{Percy Liang}$^{2}$,  \textbf{Chao Zhang}$^{1}$, \textbf{Sherry Yang}$^{2}$, \textbf{Bo Dai}$^{1}$ \\
\\
$^{1}$Georgia Institute of Technology \\
$^{2}$Stanford University
}

\begin{document}

\maketitle

\begin{abstract}

While Language Models (LMs) have made significant progress in automating machine learning engineering (MLE), the acquisition of high-quality MLE training data is significantly constrained. Current MLE benchmarks suffer from low scalability and limited applicability because they rely on static, manually curated tasks, demanding extensive time and manual effort to produce.
We introduce \method, a fully automated multi-agent pipeline, to transform raw datasets into competition-style MLE challenges through an efficient \emph{generate--verify--execute} paradigm for scaling MLE tasks with verifiable quality, real-world usability, and rich diversity. 
The proposed multi-agent pipeline in \method drives structured task design and standardized refactoring, coupled with a hybrid verification mechanism that enforces strict structural rules and high-level semantic soundness. It further validates empirical solvability and real-world fidelity through interactive execution.
We apply \method to 224 of real-world datasets and generate 606 tasks spanning multiple categories, objectives, and modalities, demonstrating that \method can work 
effectively 
across a wide range of real-world datasets.
Evaluation on the generated tasks shows that the performance of eight mainstream and cutting-edge LLMs on \method tasks is strongly correlated with their performance on carefully human-designed tasks, highlighting the effectiveness of the \method to scaling up MLE tasks, while maintaining task quality.

\end{abstract}

\vspace{-0.15cm}
\section{Introduction}
\vspace{-0.15cm}
\label{sec:intro}


Large Language Model (LLM) agents have demonstrated remarkable capabilities in automating complex coding and engineering domains~\citep{chan2024mle, qiang2025mle, nathani2025mlgymnewframeworkbenchmark, jing2024dsbench, yang2024swe, jimenez2023swe}, with machine learning engineering (MLE) emerging as a key frontier for evaluating the capability of models today. The development of sophisticated MLE agents, capable of autonomously handling tasks from data pre-processing to model tuning and deployment, promises to revolutionize scientific discovery and industrial applications. However, evaluating and developing such agents poses a significant challenge, due to the inherent complexity of MLE workflows, the need for domain-specific knowledge, and the iterative, feedback-driven nature of real-world machine learning pipelines. Developing robust MLE agents, therefore, requires not only the design and implementation of agent frameworks but also the creation of holistic environments and benchmarks that support end-to-end experimentation and structured evaluation under truly real-world conditions, encompassing diverse task distributions. 

Recent efforts have established valuable benchmarks and interactive environments for evaluating and training these agents~\citep{huang2023mlagentbench, jing2024dsbench, chan2024mle, qiang2025mle, nathani2025mlgymnewframeworkbenchmark}.  
However, existing benchmarks such as MLE-Bench~\citep{chan2024mle} and DS-Bench~\citep{jing2024dsbench} and gym-like interactive environments such as MLE-Dojo~\citep{qiang2025mle} and MLGym~\citep{nathani2025mlgymnewframeworkbenchmark} offer only static collections of tasks, and their construction remains heavily reliant on extensive human curation. This manual effort stems from two main sources: (1) the competitions selected for inclusion in these benchmarks are often carefully designed by human experts, and (2) the benchmarks require substantial engineering work to adapt these competitions into a standardized format suitable for benchmarking. Such adaptation typically involves non-trivial engineering efforts such as the pre-processing and splitting of data into train and test splits, along with implementing evaluation scripts and establishing a scoring mechanism. In addition, the ambition to establish a comprehensive environment for evaluating and training MLE agents imposes further demands on the scale and diversity of available MLE tasks.
The continued reliance on static, manually curated tasks restricts the diversity and realism of interaction scenarios and introduces a scalability bottleneck that impedes the rapid development and reliable assessment of next-generation MLE agents.
Thus, overcoming this limitation necessitates an automated framework that can continuously generate, verify, and evolve MLE tasks at scale.

Building such a framework for scaling MLE tasks presents a formidable challenge: how can the framework rigorously validate the correctness and practical value of each newly generated task? Unlike conventional supervised datasets, an MLE benchmark must satisfy multiple intertwined criteria:
(i) \emph{Structural integrity}, ensuring that all associated components including data pre-processing scripts, file directory hierarchies, and evaluation pipelines, must execute end-to-end without manual intervention, ensuring that the task is reproducible and computationally viable;
(ii) \emph{Semantic soundness}, confirming that the defined learning objective must be coherent, and the input–output structure must reflect the natural affordances and signals present in the source dataset, avoiding degenerate or trivial mappings; and
(iii) \emph{Empirical solvability}, demonstrating that the task should be non-trivial yet tractable—i.e., standard baseline agents must be able to achieve meaningful performance and exhibit stable improvement under reasonable training protocols.
A failure on any of these dimensions undermines the utility of the task, preventing it from eliciting meaningful behavioral differences across agents or supporting their effective training and development in interactive settings.

To address these challenges, we present \method, a fully automated framework that transforms raw datasets into competition-style MLE tasks through a scalable \emph{generate–verify–execute} pipeline. 
\method is carefully designed to enforce structural integrity, semantic soundness, and empirical solvability by integrating a {\bf multi-agent generation workflow}, a robust hybrid verification mechanism, and an execution-based validation loop, as illustrated in Figure~\ref{fig:main}, which provides an overview of the end-to-end paradigm. The system features three specialized agents—Brainstormer, Designer, and Refactor—that generate, concretize, and standardize task proposals in a modular, auditable manner. A persistent verification mechanism, combining both deterministic checks and agent-based reviews, continuously ensures the correctness and coherence of tasks.
Finally, each task is validated by interactive execution between a validation MLE agent and MLE environments, confirming that it supports end-to-end execution and delivers non-trivial signals on the performance of ML solutions. This principled pipeline ensures that each generated task is format-consistent, executable, and verifiable, while remaining practically meaningful for training and evaluating MLE agents.

We summarize our main contributions as follows:
\begin{itemize}
\vspace{-1ex}
\item \textbf{A fully automated task generation framework.} We propose \method, the first end-to-end system that transforms raw datasets into competition-style machine learning engineering (MLE) tasks through a scalable \emph{generate--verify--execute} pipeline. Unlike prior efforts that rely on static curation, \method enables continuous generation of realistic and diverse MLE challenges at scale, without \emph{any} human intervention.
\item \textbf{A hybrid verification mechanism.} To ensure the quality and utility of generated tasks, we design a multi-layer verification mechanism that combines static format validation, semantic alignment, and execution-based tests of empirical solvability. This hybrid stack enforces rigorous guarantees on task integrity, ensuring that each constructed challenge is well-structured, executable, and grounded in realistic machine learning scenarios.
\item \textbf{A large-scale, diverse generated task suite.} We apply \method to 224 real-world datasets and produce 606 fully verified tasks spanning a wide spectrum of modalities (e.g., tabular, vision, time series), learning objectives (e.g., classification, regression, ranking), and domains (e.g., healthcare, sports). Evaluation on a representative subset of 50 tasks with eight cutting-edge LLMs reveals strong correlation with rankings of these LLMs on human-curated benchmarks, demonstrating that \method yields challenging, discriminative, and generalizable tasks suitable for evaluating and eventually training next-generation MLE agents.
\end{itemize}

\section{Related Works}
\label{sec:related_works}

\paragraph{Agent Benchmarks and Environments.}
Recent efforts have introduced a diverse suite of benchmarks and interactive environments for the evaluation and development of LLM-based agents across multiple domains, including software engineering (SWE) benchmarks~\citep{jimenez2023swe, pan2024training, yang2024swe, zhang2025swe, zan2025multi, aleithan2024swe} that test agents’ ability to modify large codebases and repair real-world bugs, web navigation and browsing tasks~\citep{chezelles2024browsergym, zhou2023webarena, pan2024webcanvas, levy2024st, wei2025browsecomp, wu2025webwalker, yao2022webshop} that evaluate agents’ capacity to navigate complex websites or device interfaces, deep research settings~\citep{du2025deepresearch, bosse2025deep, phan2025humanity} that require multi-step reasoning and information aggregation, general tool-use environments~\citep{yao2024tau, qin2023toolllm, mialon2023gaia, liu2023agentbench,luo2025mcp} that probe agents’ ability to orchestrate diverse tools and external resources, and studies of human–agent collaboration in dynamic task scenarios~\citep{shao2024collaborative}.
In the MLE domain, a growing body of testbeds assesses agents on end-to-end workflows. For example, \textsc{MLAgentBench}~\citep{huang2023mlagentbench} offers 13 curated MLE tasks with baselines and performance thresholds, \textsc{MLE-Bench}~\citep{chan2024mle} standardizes 75 Kaggle competitions for structured MLE evaluation, \textsc{DS Bench}~\citep{jing2024dsbench} includes 74 modeling tasks reflecting realistic data science processes, \textsc{MLGym}~\citep{nathani2025mlgymnewframeworkbenchmark} provides a Gym-style suite for AI research workflows, and \textsc{MLE-Dojo}~\citep{qiang2025mle} scales to over 200 fully executable MLE tasks with step-wise interaction. While these MLE platforms advance realism and breadth, they remain limited by finite, manually curated task sets. In contrast, \method proposes a fully automated framework for scalable and high-quality MLE task generation, which allows for the continual generation of novel tasks in the MLE domain.
\vspace{-1.5ex}
\paragraph{Automated Task Generation.}
Automated task generation has emerged as a promising direction for scaling agent evaluation and training. \textsc{TaskCraft}~\citep{shi2025taskcraft} creates scalable, multi-tool agentic tasks with execution traces via compositional extensions.
\textsc{AutoCodeBench}~\citep{chou2025autocodebench} generates high-difficulty, multilingual code problems with LLM-driven reverse synthesis and test validation.
\textsc{SWE-smith}~\citep{yang2025swe} synthesizes tens of thousands of bug-inducing software engineering tasks from real-world Python repositories.
\textsc{Self-Challenging}~\citep{zhou2025self} trains agents to generate and solve their own Code-as-Task problems with built-in verification, enabling high-quality self-supervised RL.
\textsc{SQLM}~\citep{chen2025self} frames task generation as asymmetric self-play, where models propose and solve increasingly challenging problems without external data.
\method serves as the first automated framework for task generation in the MLE domain, paving the way for scalable agent evaluation and training on realistic, high-quality tasks.

\vspace{-0.15cm}
\section{Methods}
\vspace{-0.15cm}
\label{sec:methods}
\begin{figure}
    \centering
    \includegraphics[width=\linewidth]{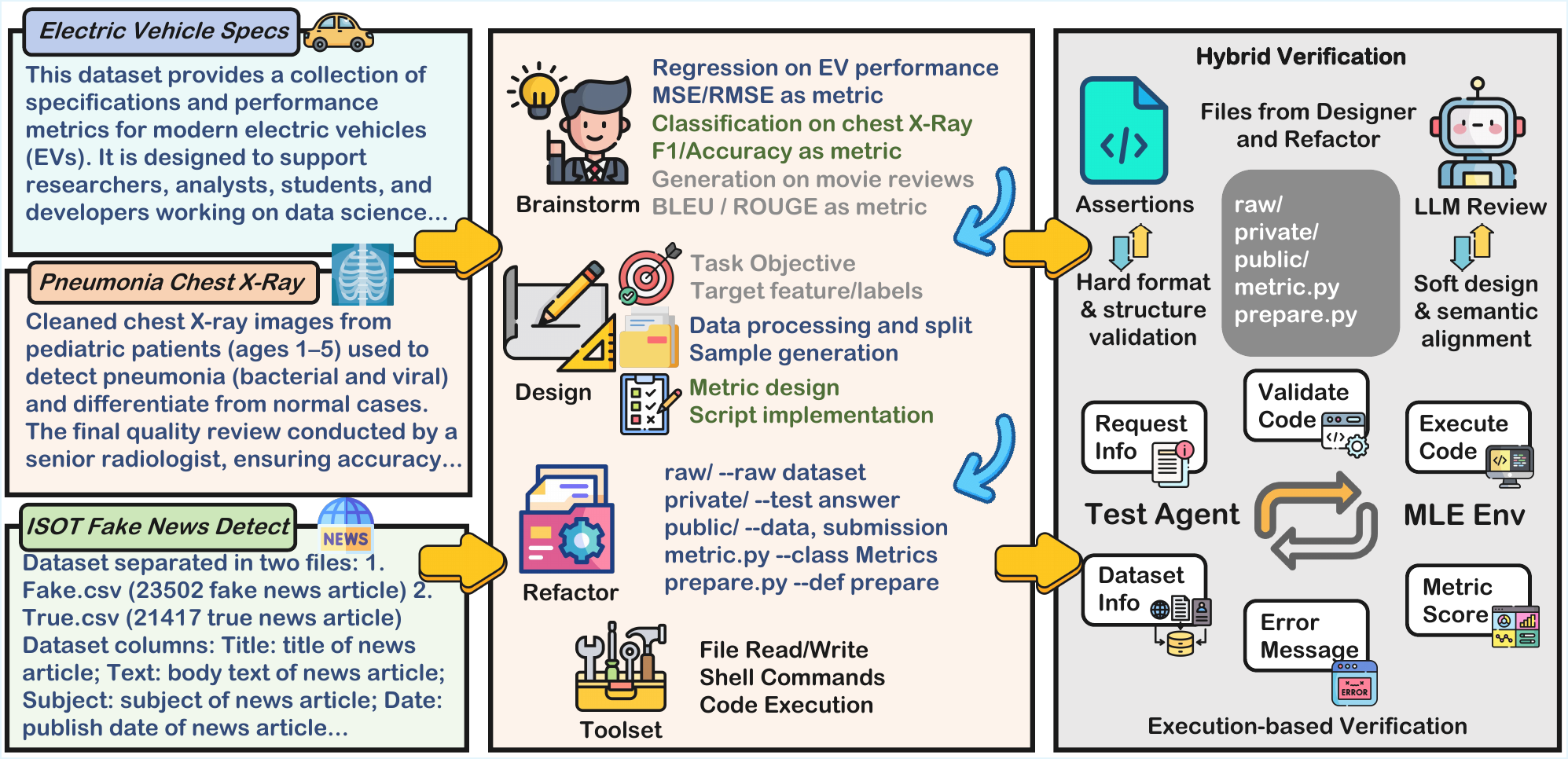}
    \caption{\method automatically generates competition-style machine learning engineering (MLE) tasks from raw datasets through a \emph{generate--verify--execute} paradigm.}
    \vspace{-3ex}
    \label{fig:main}
\end{figure}


\method automatically generates competition-style machine learning engineering (MLE) tasks from raw datasets (from sources such as Kaggle) through a \emph{generate--verify--execute} paradigm. The pipeline couples (i) \textbf{structured multi-agent generation} that designs and generates feasible tasks in multiple directions, (ii) a \textbf{hybrid verification mechanism} that enforces both hard structural constraints and soft semantic criteria, and (iii) \textbf{execution-based validation} inside an interactive MLE environment to ensure empirical solvability and real-world validity. This sequential architecture is designed to balance the diversity of task proposals with strong guarantees on the structural correctness and downstream usability of generated MLE tasks.

\vspace{-0.15cm}
\subsection{Multi-Agent Generation Workflow}
\vspace{-0.15cm}

\method employs three specialized agents that handoff generated artifacts in a sequential pipeline augmented with controlled feedback loops to allow for upstream refinement. Each agent has access to useful domain tools, including file I/O, shell commands, code execution, and always generates outputs in a pre-defined, structured format amenable to automated verification. The middle part of Figure~\ref{fig:main} illustrates how these agents sequentially advance the pipeline and produce the corresponding deliverables.


\paragraph{Brainstormer.}
Given a dataset overview along with the toolset for in-depth, multi-round data exploration, the Brainstormer enumerates a set of candidate task formulations rather than a single design, recognizing that a single dataset often supports multiple plausible learning objectives and modeling strategies. This diversity-aware generation allows the system to fully exploit the dataset’s potential. The number of candidate tasks is adaptively determined by the Brainstormer based on the dataset's intrinsic properties and structural characteristics. A key principle is that all labels and features must be accurate and grounded in the data itself, either explicitly provided or deterministically derived, rather than synthetic or heuristically constructed. 
Each proposal specifies candidate \textbf{prediction targets} (classification labels, regression variables, sequence outputs), \textbf{evaluation metrics} (e.g., accuracy, macro-F1, RMSE, or domain-specific scores), \textbf{data utilization} (e.g., preprocessing, feature construction, label extraction)  and \textbf{justifications} that articulate the rationale and practical usability of the proposed design. 
Equipped with domain tools, the Brainstormer gains comprehensive and in-depth insights, enabling it to generate grounded and valuable task proposals.
By explicitly separating hypothesis generation from commitment, \method preserves design optionality and encourages diversity without sacrificing feasibility.

\paragraph{Designer.}
For each candidate task formulation, the Designer is responsible for instantiating a fully specified machine learning engineering (MLE) task that can be executed end-to-end without manual intervention. This includes constructing 4 components necessary to define, prepare, and evaluate the task in a reproducible and verifiable manner: (i) preprocessing the raw dataset and producing deterministic training and test splits with appropriate label coverage and data integrity guarantees;
(ii) defining input and output schemas that govern the structure of model predictions and evaluation targets;
(iii) specifying the evaluation protocol and instantiating a fair, task-specific metric that captures performance with numerical stability; and
(iv) generating the complete suite of auxiliary components, including task descriptions that summarize the problem setup, data usage, and evaluation strategy; preparation scripts that performs data preprocessing, splitting, and validation checks; structured sample submission files with randomized and valid predictions; evaluation scripts for submission format validation and metric score calculation; and testing scripts to verify the correctness and consistency of the generated scripts.

Together with the original dataset, these artifacts form a complete, self-contained MLE task package that can be executed, evaluated, and iterated upon by agents in an interactive environment. Generating multiple such packages in parallel allows for efficient exploration of diverse task designs and principled comparisons across candidate formulations.

\paragraph{Refactor.}
The Refactor module standardizes all candidate task designs into a unified and well-specified format. We present the details of this structural task format in Appendix~\ref{app:unified-format}.
Rather than merely cleaning code or reorganizing files, this stage rewrites each task into a shared, consistent schema that defines the preparation interface, input/output specifications, metric implementation, canonical file structure, and feedback reporting mechanism. We define a set of conventions that govern the structure and semantics of valid tasks paired with verification routines that check conformance to these standards. By enforcing these common conventions while preserving task-specific logic, the Refactor ensures format consistency, cross-file coherence, and reliable execution. This unified representation enables downstream validation of structural correctness, streamlining automated testing pipelines to verify whether each task executes end-to-end without intervention.

\vspace{-0.15cm}
\subsection{Hybrid Verification Mechanism}
\vspace{-0.15cm}
To guarantee that every generated task is not only correct in terms of format but also semantically coherent and practically solvable, we implement a persistent \emph{Hybrid Verification Mechanism}—a multi-layered, multi-agent collaborative contract through the entire \emph{generate--verify--execute} pipeline. This mechanism executes across stages and comprises three complementary verification strategies: deterministic \textit{Assertions}, model-mediated \textit{Reviews}, and empirical \textit{Execution-based Validation}.

\vspace{-1.5ex}

\paragraph{Assertions (deterministic guards).}
Assertions encode mandatory structural constraints that are enforced through deterministic checks. These include validation of existing files, directory layout, and compliance with a structured schema for functions, classes, and scripts. Crucially, each assertion stage serves as a gatekeeper, ensuring that downstream modules can operate reliably without encountering missing inputs or malformed artifacts. Prior to Refactor, Assertions confirm the completeness and structural integrity of outputs from the Designer. As a representative example, Pre-Refactor Assertions may verify that the \emph{metric.py} and \emph{prepare.py} scripts execute correctly, and that both a \emph{sample submission} and a corresponding \emph{test answer} are successfully created. Post-Refactor, Assertions enforce full conformance to the unified task schema, including function signatures, interface formats, and execution scripts. For instance, they may examine whether the entire directory satisfies the pre-defined, unified format as in Appendix~\ref{app:unified-format}.
These rigid checks not only eliminate syntactic and structural defects but also ensure that the task satisfies all requirements for automated downstream execution. A task that successfully passes all assertions can be regarded as a fully structured and automation-ready MLE task, capable of running end-to-end without human intervention.

\vspace{-1.5ex}
\paragraph{Reviews (semantic validation).}
Where assertions enforce formal correctness, Reviews evaluate the semantic quality and intent alignment of each task. Leveraging an LLM-based agent as the reviewer, this stage assesses the clarity of task descriptions, the appropriateness of metrics, and whether the setup encourages meaningful agent behavior over shortcut solutions. For example, Reviews may flag task descriptions that omit necessary information, or ones that leak ground truths, which would pass assertions but compromise semantic validity. Though non-deterministic, Reviews serve as a soft but crucial layer that guides refinement when rigid rules are insufficient.

\vspace{-1.5ex}
\paragraph{Execution-based validation (empirical tractability).}
Beyond structural and semantic checks, a well-posed MLE task must also demonstrate empirical viability: it should admit learnable patterns, enable meaningful performance differentials, and support full-pipeline execution under realistic agentic interactions. To verify this, we introduce \emph{execution-based validation stage} that runs the entire task within an interactive MLE environment. This stage leverages a coding agent with action budgets to simulate a typical MLE agent interaction process. The environment, based on MLE-Dojo~\citep{qiang2025mle}, exposes an API for retrieving task metadata, validating code, executing scripts, and evaluating submissions. This interface allows for transparency over the actions that the step-wise agent takes and provides fine-grained feedback on execution results and performance.

The environment monitors two key aspects of empirical validation: (\textit{i}) \emph{realistic pipeline validation}, which ensures that the full pipeline, including data preparation, model training, evaluation and scoring, executes successfully without human assistance; and (\textit{ii}) \emph{performance validation}, which verifies that test agents achieve non-trivial predictive performance and that the evaluation metric exhibits sensitivity to method quality. Failures along either dimension are logged as structured defects and routed back into the verification mechanism, triggering either targeted refinement by the Refactor or Designer module or a re-execution of the corresponding stage. Positioned at the end of the generation pipeline, execution-based validation ensures empirical solvability by running the full task pipeline and measuring non-trivial agent performance. It captures failure modes that escape earlier static or semantic checks, serving as the ultimate safeguard for real-world usability.

\noindent Taken together, the three layers of verification offer distinct but complementary guarantees: \textit{Assertions} ensure structural correctness, \textit{Reviews} ensure semantic alignment, and \textit{Execution} ensures real-world solvability and usability. Only tasks that satisfy all three criteria are retained as verified, high-quality MLE challenges suitable for automated benchmarking and agent development.

\vspace{-0.15cm}
\section{Automated Task Generation}
\vspace{-0.15cm}
\label{experiments}

\method can operate seamlessly across datasets of diverse modalities, formats, and domains. To comprehensively evaluate the performance and capabilities of \method, we collect datasets from Kaggle, the most large-scale platform that hosts diverse, real-world machine-learning competitions and data resources.
We sampled 300 datasets from those with high usability scores as the experimental corpus and generated 807 tasks from these 300 source datasets.
We reserve a subset of 50 generated tasks to evaluate the quality of \method, by measuring the alignment of the performance of mainstream LLMs with the MLE-Dojo leaderboard. 

\vspace{-0.15cm}
\subsection{Agent and Environment Setups}
\vspace{-0.15cm}
We use GPT-5~\citep{gpt-5} to serve as the backbone model for all of the agents in \method. We use a default temperature of 1.0 for GPT-5. We emphasize that the proposed multi-agent pipeline is compatible with any LLM. 
For each dataset, the Brainstormer Agent is allowed up to 30 steps of tool-call actions. 
Additionally, for each source dataset, the Brainstormer Agent is allowed to brainstorm at most 3 candidate task formulations. Then, for each candidate, both the Designer and Refactor Agents have at most 3 retry times to pass all assertions.
For every proposed task formulation, the Designer and Refactor are additionally allocated a separate budget of up to 30 steps to complete their respective processes.
For the execution-based validation stage, we adapt MLE-Dojo and set up an interactive MLE environment with \texttt{request\_info} and \texttt{execute\_code} interfaces, which respectively support retrieving task-related information and evaluating submissions. The environment provides step-wise, structured feedback to agents.  
We implement a ReAct-style MLE Agent~\citep{yao2023react, sun2023adaplanner} with a budget of up to 10 steps to generate, debug, and execute code submissions to get valid metric scores.

\vspace{-0.15cm}
\subsection{Statistics of Generated Tasks}
\vspace{-0.15cm}
\begin{figure}
    \centering
    \includegraphics[width=\linewidth]{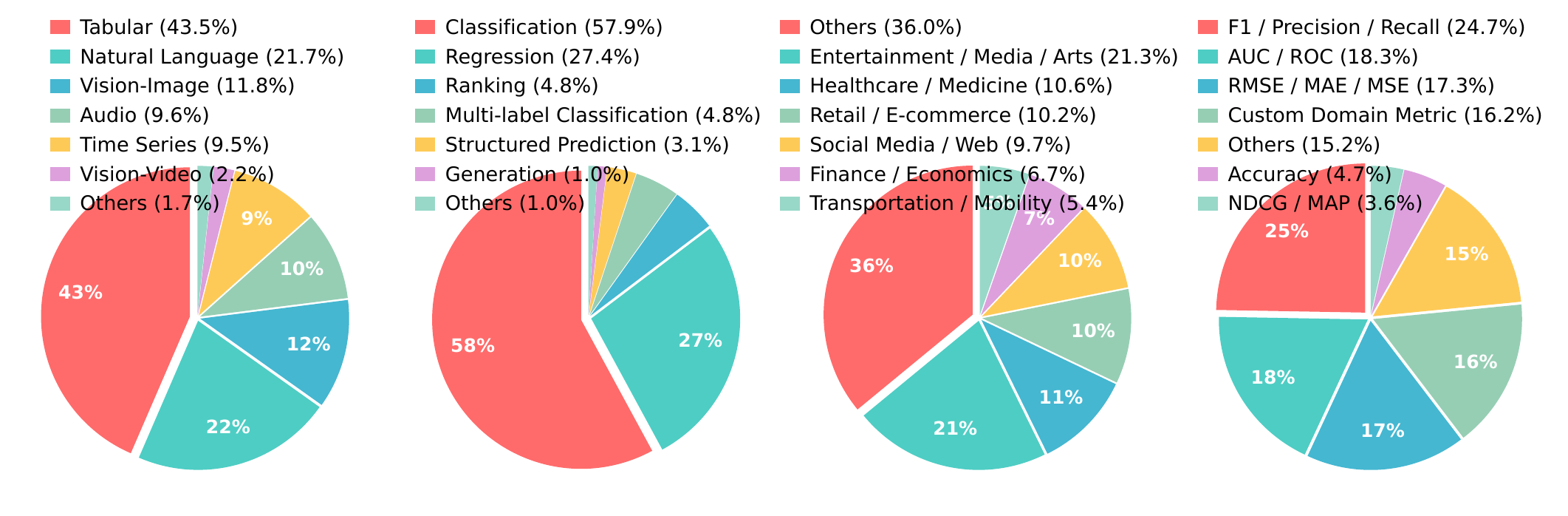}
    \vspace{-5.5ex}
    \caption{Domain, Modality, and Formulation Distribution of \method generated tasks. From left to right, the panels show the distributions of modality, objective, domain, and metric, respectively. "Others" category aggregates all types whose individual proportions are relatively minor.}
    \vspace{-3ex}
    \label{fig:diversity}
\end{figure}
\paragraph{Scale and Cost.}

\method produced a total of 606 fully verified tasks across 224 distinct source datasets, demonstrating both scalability and efficiency of our proposed approach.
On average, each dataset yields 2.71 competition-style tasks with the end-to-end preparation time per task averaging 419.98 seconds, and per dataset averaging 1136.20 seconds. This runtime excludes the execution-based verification stage, as this stage depends heavily on dataset/task characteristics, hardware configuration (GPU \& CPU), and the diversity of agent-generated code, exhibiting large variance. Here, the per-task execution time is typically below 600~seconds.
The overall pipeline incurred an average cost of \$0.78 per task and \$2.11 per dataset, including all the generation workflow and verification stages.
This time required for automatic task generation is substantially lower than the manual cost of human experts authoring competition-style tasks, and also significantly less than the engineering effort needed to localize and standardize Kaggle competitions into benchmark-ready formats. Moreover, the execution-based verification stage is negligible when compared to the time it would take for human practitioners to solve a task and achieve a meaningful score. This considerable efficiency in time strongly underscores the scalability of \method for large-scale machine learning engineering (MLE) task generation.

\vspace{-1.5ex}
\paragraph{Domain, Modality, and Formulation Diversity.}
The generated tasks span a broad spectrum of real-world data modalities, target objectives, task domains and evaluation metrics.
Figure~\ref{fig:diversity} illustrates the detailed distributions of generated tasks in these four aspects.
Specifically, the task modalities of \method generated tasks includes Tabular, Image, Video, Audio, Natural Language, Time Series, and other structured sources. Due to the characteristics of the source Kaggle datasets, tabular and natural language modalities appear more frequently. However, other modalities also constitute a substantial portion of the generated tasks. The benchmark covers a variety of formulations: while classification and regression are relatively common, it also includes ranking, multi-label classification, structured prediction, and generation tasks, offering diverse challenges for MLE agents. Compared to modality and objective, metric design tends to exhibit greater flexibility, as it is not necessarily tied to the intrinsic properties of the dataset. Thus, \method naturally reflects this flexibility. The benchmark employs a wide range of evaluation metrics, with F1, precision, and recall collectively accounting for 24.7\%, followed by AUC/ROC (18.3\%), RMSE/MAE/MSE (17.3\%), and a notable portion of custom domain-specific metrics (16.2\%). Other metrics, such as ranking-based measures like NDCG and MAP (3.6\%), further contribute to the overall diversity, highlighting the pipeline’s ability to support nuanced evaluation tailored to different task types.
\vspace{-1.5ex}
\paragraph{Agent-Wise Performance.} 
For each candidate formulation proposed by the Brainstormer, both the Designer and Refactor components are allowed up to three retries, with a maximum step limit imposed for each attempt. For different datasets and formulations, the number of retries and steps used by the Designer and Refactor components is summarized by the following statistics. In over 99\% of cases, the Designer succeeds on the first attempt and passes all assertion checks. Approximately 92\% of the time, it completes the task in no more than 15 steps, with the shortest successful case requiring only 8 steps, and none exceeding 26 steps.
In contrast, the Refactor component requires more retries and tends to take more steps: around 6\% of tasks are only completed successfully on the second attempt, and about 1\% require a third. Across all tasks and formulations, Refactor consistently uses more than 13 steps, with the majority of tasks densely utilizing 15 to 22 steps. These results align with the intended roles and design of the agents: the Refactor typically requires more actions than the Designer, as it must read the provided examples, analyze how to standardize the code and file structure to meet the required specifications, and ultimately ensure all tests pass.



\vspace{-0.15cm}
\section{Experiments: Task Evaluation}
\vspace{-0.15cm}

We evaluate whether the tasks generated by \method faithfully reflect the difficulty and discriminative structure of real, human-designed tasks. 
We conduct a comprehensive evaluation of eight cutting-edge large language models (LLMs) on a curated benchmark of 100 machine learning engineering (MLE) tasks, which we refer to as the \textbf{Combined set}. This evaluation suite comprises 50 tasks from the original MLE-Dojo evaluation set \textbf{Dojo set} and 50 tasks automatically generated by \method \textbf{Smith set}.
Both subsets are designed to span a diverse range of data modalities, application domains, and task formulations, providing a sufficiently diverse MLE testbed.

\vspace{-0.15cm}
\subsection{Experiment Setups}
\vspace{-0.15cm}
\begin{table*}[t]
\centering
\caption{Elo ratings of eight LLMs across different categories on the Dojo set, Smith set, and Combined set. For all columns, higher scores indicate better performance. The highest score in each category is highlighted in bold, and odd-numbered rows are shaded for visual clarity. }
\label{tab:mle-elo-ratings}
\renewcommand{\arraystretch}{1.15}
\resizebox{\textwidth}{!}{%
\begin{tabular}{@{}lccccccccccc@{}}
\toprule
\multirow{2}{*}{\textbf{Model}} & \multicolumn{5}{c}{\textbf{MLE-Dojo}} & \multicolumn{5}{c}{\textbf{MLE-Smith}} & \textbf{MLE-All} \\
\cmidrule(lr){2-6} \cmidrule(lr){7-11} \cmidrule(lr){12-12}
 & \textbf{MLE-Lite} & \textbf{Tabular} & \textbf{NLP} & \textbf{Vision} & \textbf{Overall} & \textbf{Vision} & \textbf{NLP/Tab.} & \textbf{Audio} & \textbf{Video} & \textbf{Overall} & \textbf{Combined} \\
\midrule
\rowcolor{gray!10}
Gemini-2.5-Pro   & \textbf{1272.0} & \textbf{1187.8} & \textbf{1303.6} & \textbf{1320.7} & \textbf{1254.6} & \textbf{1346.9} & 1000.7 & \textbf{1318.7} & \textbf{1484.1} & \textbf{1179.7} & \textbf{1214.3} \\
Gemini-2.5-Flash & 1189.7 & 1004.3 & 1254.5 & 1194.8 & 1146.7 & 1202.5 & 1009.1 & 1142.3 & 963.5 & 1079.3 & 1111.3 \\
\rowcolor{gray!10}
o4-mini          & 1019.9 & 1013.8 & 1173.2 & 1194.8 & 1068.0 & 1075.6 & \textbf{1083.5} & 1168.0 & 1114.6 & 1097.6 & 1082.9 \\
DeepSeek-Reasoner  & 1095.6 & 1101.0 & 915.7  & 1122.5 & 1064.8 & 1243.8 & 1028.9 & 1030.6 & 963.5 & 1059.1 & 1061.8 \\
\rowcolor{gray!10}
o3-mini          & 1017.3 & 1004.3 & 1004.6 & 1043.6 & 1011.9 & 1007.1 & 1017.6 & 984.7  & 936.7 & 1003.3 & 1007.6 \\
DeepSeek-Chat  & 975.4  & 976.0  & 1024.7 & 1037.4 & 990.7 & 956.2  & 1066.0 & 1055.3 & 999.5 & 1030.2 & 1011.2 \\
\rowcolor{gray!10}
GPT-4o               & 770.9  & 877.9  & 761.4  & 555.7  & 776.5 & 618.4  & 932.3  & 681.3  & 806.5 & 808.8 & 794.1 \\
GPT-4o-mini          & 659.3  & 834.9  & 562.2  & 530.5  & 686.7 & 549.5  & 861.9  & 619.0  & 731.5 & 742.0 & 716.8 \\
\bottomrule
\end{tabular}%
}
\vspace{-6ex}
\begin{flushleft}
\small
\end{flushleft}
\end{table*}
\textbf{LLMs for Evaluation.} 
We consider eight cutting-edge LLMs in the evaluation and improvement of LLMs as MLE Agents on \textbf{Combined set}. 
Specifically, we consider \texttt{gpt-4o-mini (2024-07-18)}~\citep{hurst2024gpt}, \texttt{gpt-4o (2024-11-20)}~\citep{hurst2024gpt}, \texttt{o3-mini (2025-01-31)}~\citep{o3-mini} and \texttt{o4-mini (2025-04-16)}~\citep{o4-mini} from OpenAI, \texttt{Gemini-2.5-Flash}~\citep{comanici2025gemini} and \texttt{Gemini-2.5-Pro}~\citep{comanici2025gemini} from Google, and \texttt{DeepSeek-V3.1-Chat (2025-03-24)}~\citep{deepseek-v3-1} and \texttt{DeepSeek-V3.1-Reasoner}~\citep{deepseek-v3-1} from DeepSeek as evaluation backbone LLMs.
For non-reasoning models, we set temperature=$0.0$ and top-$p=1.0$. For reasoning models, we use default model settings. We take the best performance of two runs per task per model.

\textbf{Agent and Environment Design.}
We implement the MLE Agent following the MLE-Dojo framework, which utilizes native actions and interacts with the MLE environment.
For each task and each run, the agent is allowed up to 15 action steps and a maximum of 12 hours of execution time.
The context and maximum output lengths are determined by the properties of the underlying model.

\textbf{Evaluation Metrics.} 
Each task is associated with a specific evaluation metric, which is used to compute the raw performance score for that task.
To ensure comprehensive evaluation and allow for a fair comparison across different models, we adopt \emph{Elo} ranking~\citep{chiang2024chatbot} as the primary comparative indicator. We follow Chatbot Arena~\citep{chiang2024chatbot} and estimate Elo scores by fitting a Bradley--Terry-style logistic model via maximum likelihood, using sample-weighted pairwise outcomes (wins/losses with ties treated as symmetric half-wins). We adopt a base-10 log-odds parameterization scaled to the Elo convention (scale = 400, base = 10, offset = 1000).
\vspace{-1ex}
\subsection{Main Results}

\begin{figure}
    \centering
    \includegraphics[width=\linewidth]{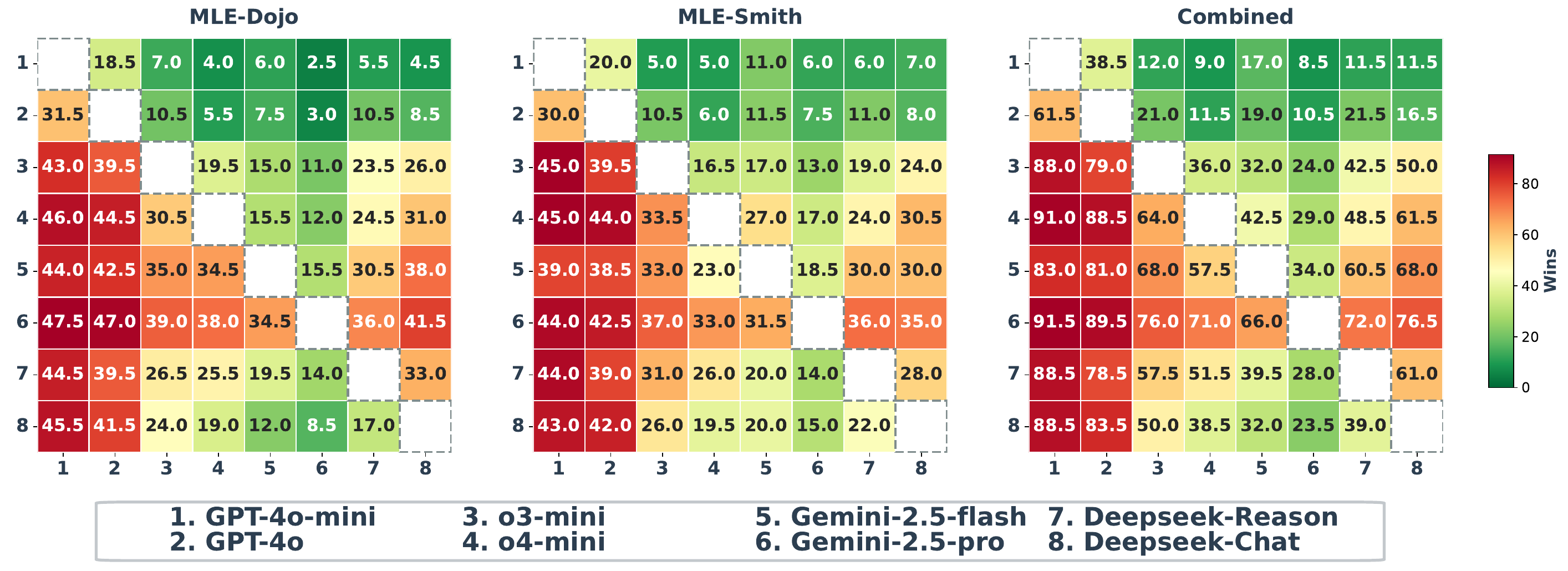}
    \vspace{-4ex}
    \caption{Pairwise win--loss matrices of eight models on the Dojo, Smith, and Combined sets.
Each cell $(i,j)$ records the number of tasks on which model $i$ outperforms model $j$,
and the aggregated score is computed by awarding $1$ point for a win, $0.5$ point for a tie, and $0$ points for a loss.}
\vspace{-3ex}
    \label{fig:win-loss}
\end{figure}
We compute modality-level Elo ratings on three disjoint sets: \textbf{Dojo set} (50 real tasks in MLE-Dojo), \textbf{Smith set} (50 \method generated tasks), and \textbf{Combined set} (all 100 tasks).
Table~\ref{tab:mle-elo-ratings} presents ELO scores for all eight LLMs across different categories and task sets.
Across all subsets, \texttt{Gemini-2.5-Pro} establishes a clear performance frontier, maintaining top rankings in almost every modality and transferring its advantage seamlessly from real to generated benchmarks. A second tier emerges with \texttt{DeepSeek-V3.1-Reasoner} and \texttt{o4-mini}, which show competitive balance across modalities: \texttt{o4-mini} is particularly strong on language-oriented tasks, while \texttt{DeepSeek-V3.1-Reasoner} delivers more robust vision performance. In contrast, the \texttt{GPT-4o} family consistently lags behind, especially on vision input, underscoring persistent challenges in multimodal generalization.
Overall, we observe a consistent ranking trend across real and synthetic tasks, validating the use of generated benchmarks for model differentiation. The Elo distribution also highlights the diversity of task difficulty and model specialization across input modalities.

\vspace{-0.15cm}
\subsection{Step-wise Performance Dynamics}
\begin{figure}
    \centering
    \includegraphics[width=\linewidth]{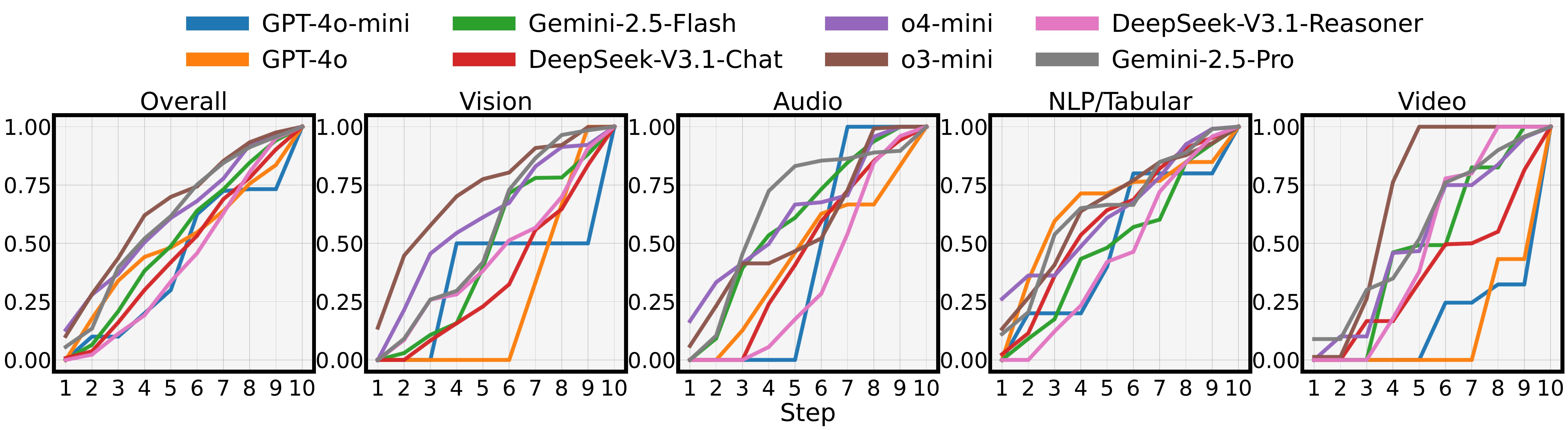}
    \vspace{-5ex}
    \caption{Step-wise Performance Dynamics of normalized raw scores. Curves are obtained by pointwise averaging over tasks in corresponding categories. Information-requesting steps are excluded.}
    \vspace{-3ex}
    \label{fig:step-dynamics}
\end{figure}

\vspace{-0.15cm}
\label{subsec:step-dynamics}
We study step-wise performance dynamics across different models to reveal consistent improvement patterns that reflect desirable properties of the automatically generated tasks.
We exclude information-requesting steps of agents and denote the remaining steps as $u\in\{1,\dots,10\}$. Since realistic leaderboards and human performances are not available for generated tasks, we implement a normalization mechanism to model step-wise improvement. For each (task $t$, model $m$), raw scores are extracted from execution feedback of \texttt{execute\_code} actions and normalized in a metric-aware manner depending on whether higher or lower values indicate better performance. Detailed formulas are provided in Appendix~\ref{app:normalization}.  
After normalization, missing entries are imputed, and we construct a best-so-far trajectory via a prefix maximum, yielding a nondecreasing length-10 curve per (task, model). Category-level and overall curves in Figure~\ref{fig:step-dynamics} are obtained by averaging across task trajectories.  
Across all categories, models exhibit consistent upward trajectories, indicating that agent performance reliably improves with steps. This trend suggests that \method-generated tasks are learnable, provide sufficient resolution to differentiate between modeling approaches, and support iterative refinement and methodical exploration. These observations provide empirical justification for using \method-generated tasks in the evaluation and development of MLE agents.

\vspace{-0.15cm}
\subsection{Realism and Quality of Generated Tasks}
\vspace{-0.15cm}
\label{subsec:realism-quality}

To evaluate the realism and discriminative fidelity of tasks generated by \method,
we analyze the statistical alignment between model-level Elo scores computed on Dojo set, Smith set, and Combined set.
Specifically, we adopt complementary statistics that capture distinct notions of agreement:
(i) \textbf{linear correlation} (Pearson~\citep{pearson1895vii}) to quantify similarity in absolute Elo magnitudes,
(ii) \textbf{rank agreement} (Spearman~\citep{spearman1961proof}, Kendall~\citep{kendall1938new}) and \textbf{head-of-leaderboard overlap} (Top-$k$) to assess stability of model ordering,
(iii) \textbf{scale and bias agreement} (Lin’s Concordance Correlation Coefficient~\citep{lawrence1989concordance}, \emph{CCC}, and Bland–Altman analysis~\citep{bland1986statistical}),
and (iv) \textbf{multi-rater reliability} (Cronbach’s $\alpha$~\citep{cronbach1951coefficient}, ICC~\citep{shrout1979intraclass}) to test whether different Elo sets function as interchangeable evaluators over the same population. We include the details of these measurements in Appendix~\ref{app:elo-metrics}.

\begin{table}[h]
\centering
\small
\vspace{-3ex}
\caption{Elo agreement with complementary statistics.
CCC denotes Lin’s concordance correlation coefficient; Kendall $\tau_b$ accounts for ties.}
\label{tab:elo-agreement}
\begin{tabular}{lcccccc}
\toprule
Pair & Pearson $r$ & $R^2$ & Spearman $\rho$ & Kendall $\tau_b$ & CCC & Top-3 / Top-5 \\
\midrule
Dojo--Smith     & 0.982 & 0.964 & 0.952 & 0.857 & 0.958 & 1.0 / 0.8 \\
Dojo--Combined  & 0.996 & 0.992 & 0.976 & 0.929 & 0.989 & 1.0 / 0.8 \\
Smith--Combined & 0.995 & 0.990 & 0.976 & 0.929 & 0.989 & 1.0 / 1.0 \\
\bottomrule
\end{tabular}
\vspace{-1.5ex}
\end{table}
Across all pairs, linear relationships remain near-perfect:
Dojo--Smith $r=0.982$,
Dojo--Combined $r=0.996$,
and Smith--Combined $r=0.995$
($R^{2}=\{0.964,0.992,0.990\}$).
Rank order is likewise stable with Spearman
$\rho=\{0.952,0.976,0.976\}$ and Kendall
$\tau_{b}=\{0.857,0.929,0.929\}$;
top rankings nearly coincide
(Top-3 overlap $=1.0$ for all, Top-5 $=\{0.8,0.8,1.0\}$).
Beyond correlation, numerical agreement is strong:
CCC $\{0.958,0.989,0.989\}$,
negligible Bland–Altman bias,
and limits of agreement of roughly
$\pm96$, $\pm51$, and $\pm45$ Elo.
Treating the three sets as interchangeable evaluators yields
$\alpha=0.993$ and $\mathrm{ICC}(2,1)=0.981$,
indicating excellent inter-set reliability.
These statistics consistently indicate that the Elo distributions induced by
\method are statistically indistinguishable from those of human–designed
benchmarks, demonstrating that \method effectively generates tasks with realistic difficulty and practical usability, faithfully mirroring the discriminative structure of real MLE competitions and supporting MLE agent development at scale.

\vspace{-0.15cm}
\section{Conclusion}
\vspace{-0.15cm}
\label{conclusion}
We introduce \method, a fully automated multi-agent pipeline for transforming raw datasets into competition-style machine learning engineering tasks. 
Through a principled \emph{generate--verify--execute} paradigm, \method scales task generation while ensuring structural integrity, semantic soundness, and empirical solvability. 
Applied to hundreds of real-world datasets, it produces a large and diverse suite of high-quality tasks that strongly correlate with human-designed benchmarks, 
demonstrating that generated tasks can match real competitions in realism and discriminative power. 



\bibliography{iclr2026_conference}
\bibliographystyle{iclr2026_conference}

\clearpage
\appendix
\section{Appendix}
\label{appnendix}



\subsection{Full List of Evaluation Tasks}
Figure~\ref{tab:eval-datasets} presents the raw dataset information of \textbf{Smith set} in dataset names, sizes and tags. The data sizes are relatively large to cover across different domains, modalities and formulations.
\renewcommand{\arraystretch}{1.1}
\setlength{\tabcolsep}{6pt}

\begin{longtable}{@{}p{0.33\textwidth} c p{0.5\textwidth}@{}}
\caption{Summary of Kaggle Competition Datasets}\label{tab:eval-datasets}\\
\toprule
\textbf{Dataset Name} & \textbf{Size} & \textbf{Tags} \\
\midrule
\endfirsthead

\toprule
\textbf{Competition Name} & \textbf{Size} & \textbf{Tags} \\
\midrule
\endhead

\bottomrule
\endfoot
\multicolumn{3}{@{}l@{}}{\textbf{Vision--General}}\\\addlinespace
veeralakrishna/200-bird-species-with-11788-images & 1.1~GB  & universities and colleges, biology, online communities \\
sadhliroomyprime/cattle-weight-detection-model-dataset-12k & 44.1~GB & animals, business, agriculture, artificial intelligence, computer vision, pre-trained model \\
muhammetzahitaydn/hardhat-vest-dataset-v3 & 4.2~GB & intermediate, deep learning, public safety, yolo, object detection \\
balraj98/modelnet40-princeton-3d-object-dataset & 1.9~GB  & earth and nature, science and technology \\
sunilthite/ovarian-cancer-classification-dataset & 3.3~GB  & cancer, pre-trained model \\
iamtapendu/rsna-pneumonia-processed-dataset & 10.9~GB & healthcare, computer vision, image, image classification, image segmentation \\
pranavchandane/scut-fbp5500-v2-facial-beauty-scores & 1.1~GB & people, computer vision, cnn, image, regression \\
majdouline20/shapenetpart-dataset & 1.0~GB & computer science, classification, segmentation \\
thedatasith/sku110k-annotations & 13.2~GB & retail and shopping \\
tapakah68/supervisely-filtered-segmentation-person-dataset & 4.3~GB & arts and entertainment, people, computer vision, image \\
aletbm/urban-segmentation-isprs & 6.4~GB & earth and nature, data visualization, classification, image classification, image segmentation \\
hendrichscullen/vehide-dataset-automatic-vehicle-damage-detection & 2.1~GB & image, multiclass classification, insurance, object detection, segmentation \\
victorcallejasf/multimodal-hate-speech & 6.0~GB & nlp, image, multiclass classification, online communities, social networks \\
\multicolumn{3}{@{}l@{}}{\textbf{Audio}}\\\addlinespace
yashdogra/speech-commands & 2.3~GB & tensorflow, automatic speech recognition, speech synthesis, speech-to-text \\
daviddkarnowski/amateur-radio-transmissions-2-meter-fm-simplex & 34.0~GB & mobile and wireless, electronics, signal processing, audio, audio classification \\
soumendraprasad/sound-of-114-species-of-birds-till-2022 & 2.1~GB & arts and entertainment, earth and nature, beginner, intermediate, advanced, audio \\
mathurinache/the-lj-speech-dataset & 3.0~GB & artificial intelligence, advanced, signal processing, text, audio \\
chrisfilo/urbansound8k & 5.6~GB & arts and entertainment, music, classification, multiclass classification, audio \\
vjcalling/speaker-recognition-audio-dataset & 7.3~GB & arts and entertainment, music, classification, deep learning, audio \\
ikrbasak/sep-28k & 2.2~GB & healthcare, health, audio, numpy, scipy \\
abdelrahmanahmed110/quran-audio-dataset & 3.0~GB & music, religion and belief systems, audio \\
raajanwankhade/oep-dataset & 11.0~GB & universities and colleges, computer vision, audio event classification, object detection, video classification \\
aryashah2k/noise-reduced-uaspeech-dysarthria-dataset & 8.0~GB & music, computer science, software, deep learning, audio synthesis, automatic speech recognition, audio classification, speech synthesis \\
jesusrequena/mlend-spoken-numerals & 1.1~GB & culture and humanities, languages, signal processing, audio \\
victorling/librispeech-clean & 28.1~GB & audio \\
imsparsh/deam-mediaeval-dataset-emotional-analysis-in-music & 1.8~GB & music, intermediate, advanced, multiclass classification, audio \\
vinayshanbhag/bird-song-data-set & 2.1~GB & music, audio \\
\multicolumn{3}{@{}l@{}}{\textbf{NLP / Tabular}}\\\addlinespace
devdope/900k-spotify & 1.0~GB & arts and entertainment, music, education, text generation \\
fayaznoor10/movie-transcripts-59k & 860.4~MB & arts and entertainment, movies and tv shows, nlp, text mining, multilabel classification \\
gowrishankarp/newspaper-text-summarization-cnn-dailymail & 503.3~MB & literature, nlp, text, news, transformers \\
nadyinky/sephora-products-and-skincare-reviews & 146.8~MB & computer science, nlp, recommender systems, retail and shopping, ratings and reviews \\
arshkon/linkedin-job-postings & 158.8~MB & employment, income, business, economics, nlp, jobs and career \\
sobhanmoosavi/us-traffic-congestions-2016-2022 & 2.3~GB & united states, categorical, transportation, tabular, urban planning \\
kgmuchiri/world-athletics-all-time-dataset & 52.9~MB & running, sports, data visualization, data analytics, tabular \\
edwardgaibor/pfaf-medical-plants-use-dataset & 13.9~MB & biology, agriculture, beginner, tabular, text \\
imoore/60k-stack-overflow-questions-with-quality-rate & 21.0~MB & music, nlp, text mining, text \\
spsayakpaul/arxiv-paper-abstracts & 44.6~MB & education, nlp, multilabel classification \\
arushchillar/disneyland-reviews & 11.1~MB & business, nlp, data visualization, tabular, ratings and reviews \\
simaanjali/emotion-analysis-based-on-text & 31.9~MB & earth and nature, nlp \\
jaidityachopra/esg-sustainability-reports-of-s-and-p-500-companies & 23.8~MB & nlp, investing, feature extraction, text pre-processing \\
smagnan/1-million-reddit-comments-from-40-subreddits & 71.2~MB & arts and entertainment, categorical, nlp, binary classification, online communities, social networks \\
salah1992/arabic-nli-pairs-multilingual-nli-26lang-2mil7 & 23.7~MB & earth and nature, linguistics, nlp, text, transformers, arabic \\
thedevastator/pubmed-article-summarization-dataset & 654.3~MB & bayesian statistics, earth and nature, nlp, text mining \\
shivamb/legal-citation-text-classification & 14.9~MB & australia, government, law, nlp, text \\
\multicolumn{3}{@{}l@{}}{\textbf{Vision--Video}}\\\addlinespace
zaber666/meld-dataset & 11.0~GB & signal processing, text mining, text, audio, pre-trained model \\
rohanmallick/kinetics-train-5per & 33.3~GB & earth and nature, computer vision, deep learning, video, audio \\
matthewjansen/ucf101-action-recognition & 6.5~GB & computer vision, deep learning, video, transfer learning, video classification \\
rohitsuresh15/radroad-anomaly-detection & 7.3~GB & law, automobiles and vehicles, image, video, eyes and vision, urban planning \\
elin75/localized-audio-visual-deepfake-dataset-lav-df & 23.1~GB & advanced, video, audio \\
saberghaderi/-dfl-bundesliga-460-mp4-videos-in-30sec-csv & 10.1~GB & football, sports, science and technology, video, simulations \\
\bottomrule
\end{longtable}

\subsection{Unified Task Structure}
\label{app:unified-format}
The \textbf{Refactor} should deliver each task as a unified task format, specifically following the below directory structure. The assertions will ensure the existence of essential files and directories such as \texttt{prepare.py}, \texttt{metric.py}, \texttt{description.txt}, \texttt{sample\_submission.csv}, \texttt{test\_answer.csv}, \texttt{raw/}, \texttt{public/} and \texttt{private/}. Furthermore, assertions will ensure that the implementations of \texttt{prepare.py} and \texttt{metric.py} strictly follow the required format.
Specifically, \texttt{prepare.py} must exactly implement a def prepare function whose input arguments include raw/, public/, and private/ directories.
Likewise, \texttt{metric.py} must exactly implement a Metric class that inherits from the designated base class and provides the corresponding methods for task-aware submission validation and metric evaluation.

\begin{figure}[h]
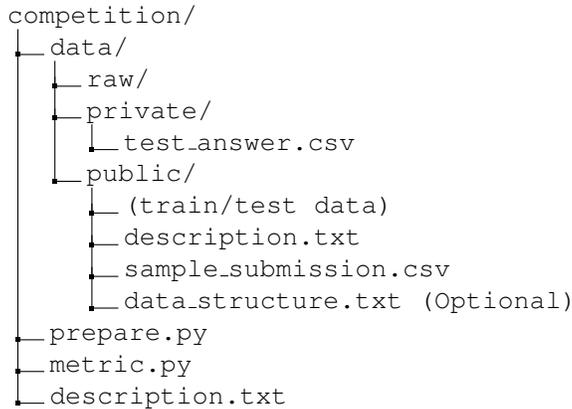


\dirtree{%
 .1 competition/.
 .2 data/.
 .3 raw/.
 .3 private/.
 .4 test\_answer.csv.
 .3 public/.
 .4 (train/test data).
 .4 description.txt.
 .4 sample\_submission.csv.
 .4 data\_structure.txt (Optional).
 .2 prepare.py.
 .2 metric.py.
 .2 description.txt.
}
\caption{Unified directory structure that \textbf{Refactor} should deliver.}
\label{fig:dir-structure}
\end{figure}

\subsection{Normalization Details}
\label{app:normalization}

For each task $t$ and model $m$, let $r_{t,m,u}$ denote the raw score from execution feedback at step $u\in\{1,\dots,10\}$.  
We define $\mathcal{D}_t\in\{+1,-1\}$ as the metric direction of task $t$, where $\mathcal{D}_t=+1$ indicates that higher metric values are better, and $\mathcal{D}_t=-1$ indicates that lower values are better.  

The normalized score is computed as:
\[
\tilde r_{t,m,u} =
\begin{cases}
\dfrac{r_{t,m,u}-\min_{u} r_{t,m,u}}{\max_{u} r_{t,m,u}-\min_{u} r_{t,m,u}},
& \mathcal{D}_t = +1,\\[8pt]
\dfrac{\max_{u} r_{t,m,u}-r_{t,m,u}}{\max_{u} r_{t,m,u}-\min_{u} r_{t,m,u}},
& \mathcal{D}_t = -1.
\end{cases}
\]

If $\max r_{t,m}=\min r_{t,m}$, observed entries are set to $1$ and missing ones to $0$. We then forward-fill missing indices and compute a best-so-far trajectory via a prefix maximum:
\[
y_{t,m,u} \;=\; \max\bigl(y_{t,m,u-1},\, \tilde r_{t,m,u}\bigr).
\]

This procedure yields a nondecreasing curve of length 10 per (task, model), which is then averaged pointwise across tasks to obtain category-level and overall trajectories.

\subsection{Prompts for \method Agents}
We provide detailed prompts for \method Agents in this section.

\vspace{2ex}
\begin{Verbatim}
You are an expert Kaggle competition designer. Your task is to 
brainstorm diverse design choices for challenging, high-quality and 
reasonable Kaggle competitions based on an existing dataset.

You are provided with detailed information about the dataset.
The dataset is already downloaded to the working directory with unzip.

You have access to tools for reading/writing files, listing directory 
structure, and executing bash commands.

Always use function calls when you need to perform actions. 
You can only call one function at a time.

Your work directory is {working_directory}, all actions and files 
should be performed in this directory.

Use tools to explore the dataset to get insights. 

Then based on insights from the dataset, brainstorm design choices for 
challenging, high-quality and reasonable Kaggle competitions.

The design choice should include the following aspects, 
be concise and informative:
- Concise problem overview: background, problem statement, and goal.
- Data utilization: for the given dataset, 
  what data to use, what data to ignore.
- Data processing: how to process the data
- Metric: what metric to use, why it's fair and precise.
- Justification of the design choices: why the designed competitiion 
  would be high-quality, challenging and solvable by ML techniques.
- Details of the ignored data: why the ignored data is not used, 
  what information is missing.
- Difficulty level: how difficult the competition is, 
  where the difficulty comes from.
- Tags: what tags the competition should be tagged with.

Principles:
- Only split the data into train and test sets.
- Ensure that only precise, reliable labels are used; no uncertain, 
  ambiguous, or model-generated labels should be introduced.
- You must brainstorm and write at least one and 
  at most {count} results. Determine the number of brainstorming outputs 
  according to the intrinsic nature and properties of the dataset.
  - Some datasets are open-ended and naturally admit a wide range of 
  tasks, while others are more specific and concentrated.
  - Aim to explore as many meaningful possibilities as the 
  data genuinely supports, but do not force artificial variety—
  respect the dataset's natural boundaries.

Write your brainstorming results in "brainstorming\_i.md" file to 
the working directory {working_directory},
is the index of the brainstorming result, from 1 to at most {count}. 


## DATASET INFORMATION ##
{dataset_information}
\end{Verbatim}

\vspace{2ex}
\begin{Verbatim}
You are an expert Kaggle competition designer. Your task is to create a 
challenging, high-quality Kaggle competition based on existing dataset.

You are provided with detailed information about the dataset. And the 
dataset is already downloaded to the working directory with unzip.

You have access to tools for reading/writing files, listing directory 
structure, and executing bash commands.

Always use function calls when you need to perform actions. 
You can only call one function at a time.

Your work directory is {working_directory}, 
all actions and files should be performed in this directory.

Now there is "brainstorming.md" file in the working directory, 
pointing out the design direction for the competition.

## REQUIREMENTS ##
- Refer to "brainstorming.md" file for the design direction.
  But follow the requirements and instructions below.
- Make the competition challenging while maintaining a high quality 
  standard.
- Participant should utilize ML techniques to solve the problem, 
  including but not limited to:
    - Data Processing
    - Feature Engineering
    - Model Training
    - Model Evaluation
    ...
- Design the metric to be reasonable, fair and precise.
- Make good use of the data as possible, don't waste any good resources. 
  Keep the scale rather than using subsets. 
  No need to care about the runtime.
- Split the data into train/test sets appropriately.
- Always specify exactly the absolute path as arguments.
- All actions and files should be performed in the working directory.
- The competition should be challenging, but solvable by ML techniques.
- Make everything perfect rather than just trying to pass the tests.

A general pipeline for reference:

1. Utilize list_directory_structure tool to explore data structure.
2. Explore data files using the read_file tool; 
   further extract files with bash commands if needed.
3. Design a concise and informative problem description 
   and write it to "description.txt" in the working directory:
    - Include the problem statement,data description,evaluation metric, 
      and any other relevant information.
    - Specify the final train/test data files for the competition, 
      while don't specify the path of the data files.
    - Ignore timeline/prize/etc, they are not needed.
4. Write a "prepare.py" file:
    - Include complete train/test split process and 
      sample_submission.csv generation
    - sample_submission.csv better has random but valid labels
      (same category as in test_answer.csv) rather than null values
    - Test_answer and test_data (without the predicted labels) should be 
      separated into two files, use "test_answer.csv" as the name
    - Consider the correspondence between the test_answer and test_data
    - Include detailed and comprehensive assert checks for the 
      correctness of the split
    - Specify the final train/test data files for the competition, 
      align with the description.txt
    - Validation set isn't needed, but keep it if it's already split
    - The image, audio, and other related files should also be split 
      together with the CSV files into train/test sets/folders.
    - Rename files with names that might reveal their labels to 
      avoid label leakage.
    - Don't include data paths in csv files
    - Set deterministic behavior for the split process.
    - For classification tasks, all test labels should occur in 
      training set at least once
5. Write a "metric.py" file, include functions to validate the format 
correctness of the submission and calculate the metric. 
Deal with numerical values carefully to avoid nan/inf/etc.
6. Write a "test.py" script to test the correctness of the prepare.py 
and metric.py, run it to check the correctness until totally correct.
7. Optimize description.txt: 
    - No need to mention the original data files, only the final data 
      files should be mentioned
    - Take the view of a participant to review it 
      (which means test_answer or irrelevant files shouldn't 
      be mentioned) and make it perfect
    - Make sure the competition is challenging, meaningful and solvable 
      by ML techniques, and the metric is fair and precise.
    - Make sure the description is informative, concise and accurate.
8. Optimize until all requirements are met with high quality
(The test must pass).


## DATASET INFORMATION ##
{dataset_information}
\end{Verbatim}

\vspace{2ex}
\begin{Verbatim}
You are an expert Python developer. Your task is to refactor several 
Python files to meet some requirements.

You have access to tools for reading/writing files, listing directory 
structure, and executing bash commands.

You are provided with the working directory: {working_directory}, all 
actions and files should be performed in this directory.

All files you need are in the working directory. raw/ is where the data 
is downloaded and unzipped once.

samples/ directory is a good example, you can refer to it first to 
learn good practices and refactor the files to meet the requirements.

You may need to check the data files for details if needed.

## REQUIREMENTS ##
- You should finally refact metric.py and prepare.py to meet the 
requirements.
- metric.py should inherit from samples/base_metric.py and implement 
the abstract methods, give it a related name that ends with "Metrics", 
refer to samples/sample_metric.py for the implementation details.
  - "evaluate" and "validate_submission" should be implemented and 
    aligned with "sample_submission.csv" and "test_answer.csv"
  - In addition to "self", "__init__" should have two arguments: 
    "value" and "higher_is_better" (Determine the default); 
    "evaluate" should have two arguments: "y_true" and "y_pred"; 
    "validate_submission" should have two arguments: 
    "submission" and "ground_truth"
- prepare.py should implement exactly "def prepare(raw: Path, 
  public: Path, private: Path)" 
  - This function is a complete preparation process
  - Refer to samples/sample_prepare.py for the implementation details
  - Set deterministic behavior for the split process.
  - test_answer (participants shouldn't see) should be placed exactly 
    in "private/" directory, other files (sample_submission, test/train 
    data/images/audio/video/text/other, etc.) should be placed exactly 
    in "public/" directory
- Write a comprehensive "test.py" script to test the correctness of 
the prepare.py and metric.py, and run it to check the correctness. 
Test "evaluate" and "validate_submission" of the metric.py with
"test_answer.csv" and "sample_submission.csv".
- Make sure the test.py is correct and comprehensive, 
and the execution of test.py is correct.
- Don't include "main" function in metric.py and prepare.py
- Always specify exactly the absolute path as arguments.
- All actions and files should be performed in the working directory.
- Finally, there should be "private/", "public/", "samples/", "raw/" 
directories, and "description.txt", "metric.py", "prepare.py", 
"test.py" files in the working directory.
  - "raw/" directory should contain the original data files
  - "private/" directory should contain the test_answer.csv file
  - "public/" directory should contain the sample_submission.csv and 
    all train/test data/images/audio/video/text/other files 
    and description.txt. There should always be "test.csv" 
    and "train.csv" in the "public/" directory if applicable. 
  - Don't include or leak anything related to answers/golden labels 
    in "public/" directory.
  - File directories in "description.txt" should be the same as the 
    exact file directories in "public/" directory. Don't mention 
    "private/" in the description.txt, only include files in "public/" 
    directory.
  - "description.txt" is open to participants, so make it concise and 
    informative, only include "public/" directory in the description.txt.
- Make everything perfect rather than just trying to pass the tests.
Optimize until all requirements are met with high quality 
(The test must pass).
\end{Verbatim}

\clearpage
\subsection{Details of Statistical Measures for Elo Set Agreement}
\label{app:elo-metrics}

This section provides formal definitions, interpretation, and common use cases
for all agreement statistics used to compare model-level Elo scores across
different task sets.

\subsubsection{Pearson Linear Correlation ($r$)}
\paragraph{Definition.}
Given paired observations $\{(x_i,y_i)\}_{i=1}^n$,
\[
r \;=\;
\frac{\displaystyle \sum_{i=1}^{n} (x_i-\bar{x})(y_i-\bar{y})}
     {\sqrt{\displaystyle \sum_{i=1}^{n}(x_i-\bar{x})^2}
      \sqrt{\displaystyle \sum_{i=1}^{n}(y_i-\bar{y})^2}} .
\]

\paragraph{Meaning.}
Measures the strength of \emph{linear} association between two sets of scores.
$r=1$ indicates perfect positive linearity, $r=0$ no linear association.

\paragraph{Use.}
Commonly used to assess whether two measurement methods produce proportionally
similar values (e.g., Elo magnitudes across task sets).

\subsubsection{Coefficient of Determination ($R^2$)}
\paragraph{Definition.}
For a simple linear regression $y_i = a + b x_i + \varepsilon_i$,
\[
R^2 \;=\; 1 - \frac{\sum_{i}(y_i-\hat{y}_i)^2}{\sum_{i}(y_i-\bar{y})^2}
      \;=\; r^2 \quad \text{(for simple correlation).}
\]

\paragraph{Meaning.}
Represents the proportion of variance in $y$ explained by $x$.
Higher $R^2$ indicates stronger predictive power of one set of scores for the other.

\paragraph{Use.}
Provides an intuitive measure of how much of the variability in Elo scores
is shared between two task sets.

\subsubsection{Spearman Rank Correlation ($\rho$)}
\paragraph{Definition.}
Let $R(x_i)$ and $R(y_i)$ be the ranks of $x_i$ and $y_i$.
\[
\rho \;=\;
\frac{\displaystyle \sum_{i}(R(x_i)-\overline{R(x)})(R(y_i)-\overline{R(y)})}
     {\sqrt{\displaystyle \sum_{i}(R(x_i)-\overline{R(x)})^2}
      \sqrt{\displaystyle \sum_{i}(R(y_i)-\overline{R(y)})^2}} .
\]

\paragraph{Meaning.}
Assesses whether the \emph{ordering} of models is preserved,
independent of absolute score scales.

\paragraph{Use.}
Robust to monotonic but nonlinear relationships,
ideal for leaderboard stability checks.

\subsubsection{Kendall Rank Correlation ($\tau_b$)}
\paragraph{Definition.}
Let $C$ be the number of concordant pairs and $D$ the number of discordant pairs.
Let $T_x$ and $T_y$ be the numbers of tied pairs in $x$ or $y$.
\[
\tau_b \;=\;
\frac{C-D}{\sqrt{(C+D+T_x)\,(C+D+T_y)}} .
\]

\paragraph{Meaning.}
Quantifies pairwise ranking agreement while properly handling ties.

\paragraph{Use.}
Often preferred when ties occur (common in Elo ratings),
providing a probabilistic interpretation:
$\tau_b$ is the difference between the probability of concordance and discordance.

\subsubsection{Top-$k$ Overlap}
\paragraph{Definition.}
For a given $k$, let $S_x^k$ and $S_y^k$ be the sets of top-$k$ ranked items:
\[
\mathrm{Overlap}_k
= \frac{|S_x^k \cap S_y^k|}{k}.
\]

\paragraph{Meaning.}
Measures how consistently the \emph{leaders} (top models) coincide.

\paragraph{Use.}
Highlights agreement in the most competitive region of leaderboards,
which is often of primary interest.

\subsubsection{Lin’s Concordance Correlation Coefficient (CCC)}
\paragraph{Definition.}
Let $\mu_x,\mu_y$ be means, $\sigma_x^2,\sigma_y^2$ variances,
and $\rho$ the Pearson correlation:
\[
\mathrm{CCC}
= \frac{2\rho \sigma_x \sigma_y}{\sigma_x^2 + \sigma_y^2 + (\mu_x - \mu_y)^2}.
\]

\paragraph{Meaning.}
Assesses both \emph{precision} (correlation) and \emph{accuracy}
(closeness to the $45^\circ$ identity line).
A value of $1$ indicates perfect agreement in both scale and location.

\paragraph{Use.}
Preferred when we need to verify numerical interchangeability
beyond simple linear association.

\subsubsection{Bland--Altman Analysis}
\paragraph{Definition.}
For each pair $(x_i,y_i)$ compute
\[
\text{Difference } d_i = x_i - y_i,\quad
\text{Mean } m_i = \frac{x_i + y_i}{2}.
\]
The plot of $d_i$ versus $m_i$ reveals systematic bias.
The \emph{limits of agreement} (LoA) are
\[
\overline{d} \pm 1.96 \, s_d ,
\]
where $\overline{d}$ is the mean difference and $s_d$ its standard deviation.

\paragraph{Meaning.}
Visualizes bias and scale discrepancies
even when correlation is high.

\paragraph{Use.}
Widely used in clinical and experimental settings
to test whether two measurement methods can be used interchangeably.

\subsubsection{Cronbach’s $\alpha$}
\paragraph{Definition.}
Suppose $k$ parallel measurements of the same quantity.
Let $\sigma_t^2$ be the variance of the total score
and $\sigma_j^2$ the variance of each measurement:
\[
\alpha \;=\;
\frac{k}{k-1}
\left[1 - \frac{\sum_{j=1}^{k}\sigma_j^2}{\sigma_t^2}\right].
\]

\paragraph{Meaning.}
Estimates internal consistency across multiple raters or measurement methods.

\paragraph{Use.}
Values above $0.9$ indicate excellent reliability,
supporting the claim that different Elo sets can be treated as
interchangeable ``raters'' of model performance.

\subsubsection{Intraclass Correlation Coefficient (ICC)}
\paragraph{Definition.}
For the two-way random, absolute-agreement, single-measure model
(denoted ICC$(2,1)$):
\[
\mathrm{ICC}(2,1)
= \frac{MS_B - MS_E}{MS_B + (k-1)MS_E + \frac{k}{n}(MS_R - MS_E)} ,
\]
where $MS_B$ is the between-target mean square,
$MS_R$ the between-rater mean square,
$MS_E$ the residual mean square,
$k$ the number of raters (here Elo sets),
and $n$ the number of targets (models).

\paragraph{Meaning.}
Captures both correlation and absolute agreement among multiple raters.

\paragraph{Use.}
A high ICC confirms that Elo scores from different sets
can be used interchangeably in downstream evaluations.

\paragraph{Summary.}
Together, these measures provide a comprehensive assessment of agreement,
covering linear association, rank stability, numerical accuracy,
and multi-rater reliability.

\end{document}